\documentclass[letterpaper, 10 pt, conference]{ieeeconf}  

\IEEEoverridecommandlockouts                              

\usepackage{times}
\usepackage{soul}
\usepackage{url}
\usepackage[hidelinks]{hyperref}
\usepackage{graphicx}
\usepackage{amsmath}
\usepackage{amssymb}
\usepackage{booktabs}
\usepackage{algorithm}
\usepackage{algorithmic}
\usepackage{multirow}
\usepackage{array}
\usepackage{xcolor}
\usepackage{cite}
\usepackage{float}
\usepackage{csquotes}

\title{\LARGE \bf
GLAD: Grounded Layered Autonomous Driving \\
for Complex Service Tasks
}

\definecolor{bleudefrance}{rgb}{0.19, 0.55, 0.91}

\author{Yan Ding, Cheng Cui, Xiaohan Zhang, and Shiqi Zhang
\thanks{The authors are with the Department of Computer Science, SUNY Binghamton, Binghamton NY 13902.
\texttt{\{yding25, ccui7, xzhan244, zhangs\}@binghamton.edu}}%
}

\begin{document}
\maketitle
\thispagestyle{empty}
\pagestyle{empty}


\begin{abstract}
Given the current point-to-point navigation capabilities of autonomous vehicles, researchers are looking into complex service requests that require the vehicles to visit multiple points of interest. 
In this paper, we develop a layered planning framework, called GLAD, for complex service requests in autonomous urban driving.
There are three layers for service-level, behavior-level, and motion-level planning.
The layered framework is unique in its tight coupling, where the different layers communicate user preferences, safety estimates, and motion costs for system optimization. 
GLAD is visually grounded by perceptual learning from a dataset of $13.8k$ instances collected from driving behaviors. 
GLAD enables autonomous vehicles to efficiently and safely fulfill complex service requests.
Experimental results from abstract and full simulation show that our system outperforms a few competitive baselines from the literature.
\end{abstract}

\section{Introduction}

Self-driving cars are changing people's everyday lives. 
Narrowly defined autonomous driving technology is concerned with point-to-point navigation and obstacle avoidance~\cite{buehler20072005}, where recent advances in perception and machine learning have made significant achievements. 
In this paper, we are concerned with urban driving scenarios, where vehicles must follow traffic rules and social norms to perform driving behaviors, such as merging lanes and parking on the right.
At the same time, the vehicles need to fulfill service requests from end users. 
Consider the following scenario: 

\vspace{.5em}
\begin{displayquote}\emph{Emma asks her autonomous car to drive her home after work.
On her way home, Emma needs to pick up her kid Lucas from school, stop at a gas station, and visit a grocery store.
In rush hour, driving in some areas can be difficult. 
Lucas does not like the gas smell, but he likes shopping with Emma.
}
\end{displayquote}
\vspace{.5em}

The goal of Emma's autonomous car is to efficiently and safely fulfill her requests while respecting the preferences of Emma (and her family). 
We say a service request is \emph{complex}, if fulfilling it requires the vehicle to visit two or more points of interest (POIs), such as a gas station and a grocery store, each corresponding to a driving task.
Facing such a service request, a straightforward idea is to first sequence the driving tasks of visiting different POIs, and then perform behavioral and motion planning to complete those tasks.
However, this idea is less effective in practice, because of the unforeseen execution-time dynamism of traffic conditions. 
For instance, the vehicle might find it difficult to merge right and park at a gas station because of unanticipated heavy traffic. 
This observation motivates this work that leverages visual perception to bridge the communication gap between different decision-making layers for urban driving. 

\begin{figure*}
\centering
\vspace{1em}
\includegraphics[width=1.9\columnwidth]{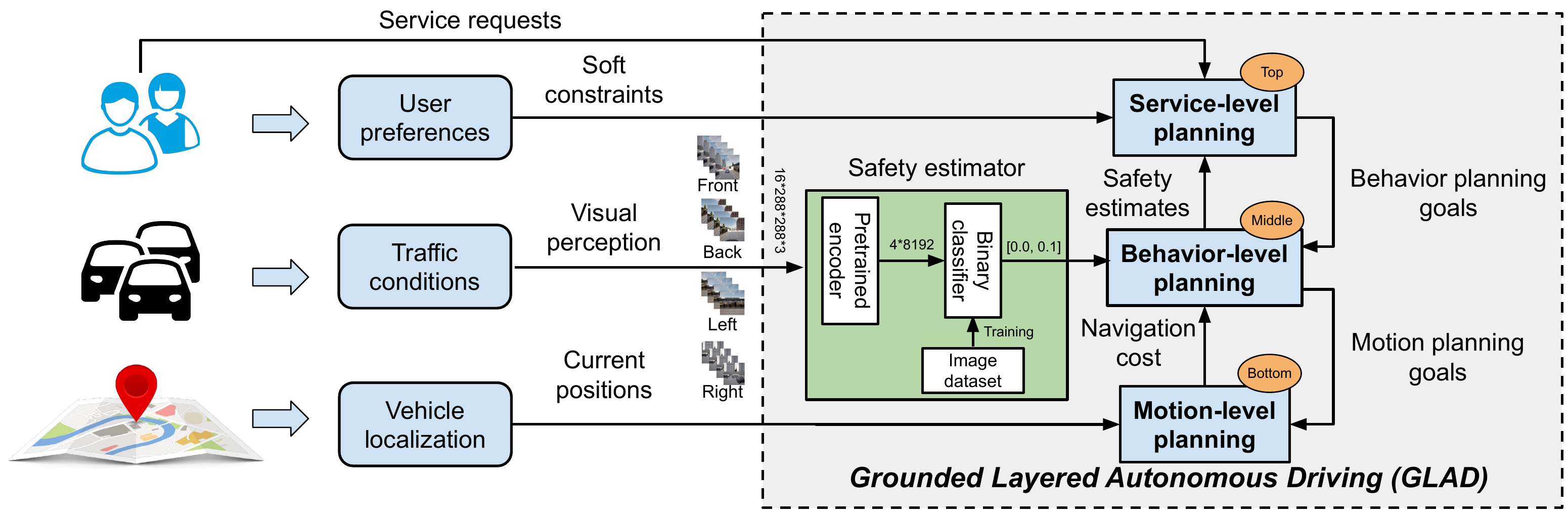}
\caption{An overview of the GLAD planning framework for complex driving tasks in urban scenarios.
GLAD consists of three decision-making layers about fulfilling service requests, sequencing driving behaviors, and computing motion trajectories respectively.
GLAD is a visually grounded planning framework, because the safety levels of driving behaviors are evaluated using computer vision.}
\label{fig:overview}
\end{figure*}

In this paper, we develop Grounded Layered Autonomous Driving (\textbf{GLAD}), a planning framework for urban driving that includes three decision-making layers for service, behavior, and motion respectively. 
The service  (\textbf{top}) layer is for sequencing POIs to be visited in order to fulfill users' service requests.
User preferences, such as ``\emph{Lucas likes shopping with Emma}'', can be incorporated into this layer.
The behavior  (\textbf{middle}) layer plans driving behaviors, such as ``merge left'' and ``drive straight''. 
The motion  (\textbf{bottom})  layer aims to compute trajectories and follow them to realize the middle-layer behaviors. 
GLAD is novel in its bidirectional communication mechanism between different layers. 
For example, the bottom layer reports motion cost estimates up to the top two layers for plan optimization. 
The safety estimates of different driving behaviors (middle layer) are reported to the top layer, and the safety estimation is conditioned on the motion trajectories in the bottom layer. 
An overview of GLAD is presented in Fig.~\ref{fig:overview}. 

``Grounding'' is a concept that was initially developed in the literature of symbolic reasoning~\cite{harnad1990symbol}. 
In this work, the vehicle's behavioral planner (middle layer) relies on symbolic rules, such as ``\emph{If the current situation is safe, a merge left behavior will move a vehicle to the lane on its left.}'' 
While classical planning methods assume perfect information about ``X is safe,'' an autonomous vehicle needs its perception algorithms to visually ground such symbols in the real world. 
We used the CARLA simulator~\cite{dosovitskiy2017carla} to collect a dataset of $13.8k$ instances, each including 16 images, for evaluating the safety levels of driving behaviors. 
Learning from our gathered dataset enables GLAD to visually ground symbolic predicates for planning driving behaviors. 
We have compared GLAD with baseline methods~\cite{chen2016combining,ding2020task} that support decision-making at behavioral and motion levels.
Results show that GLAD produced the highest overall utility compared to the baseline methods. 


\section{Background and Related Work}

Service agents sometimes need more than one action to fulfill service requests. 
\textit{Task planning} methods aim to sequence symbolic actions to complete such complex tasks. 
There are at least two types of task planning, namely automated planning~\cite{ghallab2016automated,haslum2019introduction} and planning under uncertainty~\cite{puterman2014markov}, that can be distinguished based on their assumptions about the determinism of action outcomes. 
\textit{Automated planning} systems compute a sequence of actions that lead transitions to a goal state. 
Assuming non-deterministic action outcomes, \textit{planning under uncertainty} algorithms model state transitions using Markov decision processes~(MDPs) and compute a policy that maps the current state to action. 
Other than task planning, autonomous vehicles need \textit{motion planning} algorithms~\cite{kavraki1996probabilistic,lavalle2001rapidly} to compute motion trajectories and generate control signals to follow those trajectories. 
This work involves both task planning and motion planning. 

Next, we discuss how those different planning paradigms have been used in driving and robotics literature. 

\subsection{Planning for Autonomous Driving}
Within the context of planning in autonomous driving, the majority of research has been focusing on computing trajectories connecting a vehicle's current position to its desired goal~\cite{hu2018dynamic,lim2018hierarchical, chen2019autonomous,huang2019online,zhang2020novel}.
For instance, Hu et al.~(2018) developed an approach that generates an optimal path in real time as well as the appropriate acceleration and speed for the autonomous vehicle in order to avoid both static and moving obstacles~\cite{hu2018dynamic}.
Those works concentrated on the motion planning and control problems in autonomous driving, whereas we further include task-level planning for driving behaviors and fulfilling service requests.

Driving behaviors, such as merging lanes and turning, have been modeled as a set of symbolic actions~\cite{wei2014behavioral,chen2015task,hang2020human,muzahid2021optimal}. 
Those works focused on task-level, discrete-space behavioral planning, and did not consider safety, cost, or both at the motion level. 
Another difference is that those works did not consider user preferences, whereas we do in this work. 


Recent research has incorporated state estimation into behavioral planning in autonomous urban driving. 
For instance, the work of Phiquepal and Toussaint (2019) modeled urban driving as a partially observable Markov decision process to enable active information gathering for decision-making~\cite {phiquepal2019combined}.
Another example is the work of Ding et al. (2020) that quantitatively estimated safety levels for behavioral planning~\cite{ding2020task}. 
In line with those methods, we consider partially observable worlds, and compute plans under uncertainty. 
Beyond that, GLAD has a perception component, and leverages egocentric vision for safety evaluation, which improves the real-world applicability of our approach compared with those methods. 

\subsection{Task and Motion Planning in Robotics}
Researchers in robotics have developed a variety of task and motion planning (TAMP) algorithms to enable robots to plan to fulfill task-level goals while maintaining motion-level feasibility, as summarized in review articles~\cite{garrett2021integrated,lagriffoul2018platform}. 
The majority of TAMP research concentrated on manipulation domains, where ensuring plan feasibility is key, and efficiency in task completion is secondary~\cite{toussaint2015logic,garrett2018ffrob,garrett2020online,dantam2018incremental}.
Urban driving tasks tend to be time-consuming, and sub-optimal plans might result in a very long execution time. 
Different from most TAMP methods, our work incorporates task-completion efficiency in addition to plan feasibility.

There are a few TAMP works that incorporated efficiency into plan optimization~\cite{zhang2016co,lo2020petlon,zhang2022visually}. 
For instance, the work of Zhang et al. (2022) leveraged computer vision towards computing efficient and feasible task-motion plans for a mobile manipulator in indoor environments~\cite{zhang2022visually}. 
In line with those methods, we consider both task-completion efficiency, and plan feasibility. 
Different from those TAMP methods for indoor scenarios, GLAD focuses on urban driving, and further considers user preferences and road safety. 



\section{Problem Statement}\label{sec:problem}

\noindent\textbf{Planning-time Input:}
A service request is specified as hard constraints that can be satisfied using multiple driving behaviors, denoted as $\textit{cst}^\textit{hard}$.
User preferences are formulated using soft constraints, denoted as $\textit{cst}^\textit{soft}$.
Violating such a soft constraint introduces a penalty. 
In our domains, service task specifications and user preferences are provided as part of the problem definition. 
The hard and soft constraints can be encoded in different ways depending on the syntax of planning languages and systems.

\vspace{0.8em}
\noindent\textbf{Execution-time sensory Input:}
The vehicle is provided with a 2D map, and can obtain its current position with respect to the map. 
In addition, the vehicle is equipped with four monocular cameras mounted on its front, back, left, and right sides.
Each observation \textit{IM} is in the form of a sequence of four frames, and each frame includes four images captured from the four cameras installed on the four sides of the vehicle. 
The size of \textit{IM} is $4 \times 4\times 288 \times 288 \times 3$, where $288 \times 288 \times 3$ is the image size.



\vspace{0.8em}
\noindent\textbf{Requirements:}
The driving agent is given a \textit{task planning system}, denoted as $\textit{Plnr}^t$, that can be used for sequencing driving behaviors to complete a driving task (e.g., visiting a grocery store). 
$\textit{Plnr}^t$ includes seven symbolic driving behaviors, including \texttt{mergeleft}, \texttt{mergeright}, \texttt{turnleft}, \texttt{turnright}, \texttt{gostraight}, \texttt{park}, and \texttt{stop}.
Each action (corresponding to a driving behavior) is defined by a description of its preconditions and effects. 
In this work, $\textit{Plnr}^t$ is constructed in Answer Set Programming (ASP)~\cite{lifschitz2002answer}.

The driving agent is also given a 2D \textit{motion planning system}, denoted as $\textit{Plnr}^m$.
The motion planning system can be used for computing collision-free, shortest trajectories that connect the vehicle's current and goal positions. 
The provided 2D map contains a set of POIs. 
Each POI is comprised of its name and a 2D position that can be visited by the vehicle.

\vspace{0.8em}
\noindent\textbf{Utility:}
Given a task-motion plan, the utility function considers motion costs, safety of driving behaviors, and user preferences, which are denoted as \textit{Cost}$()$, \textit{Safe}$()$, and \textit{Pref}$()$, respectively. 
Specifically, \textit{Cost}$()$ is calculated based on the vehicle's traveling time.
\textit{Safe}$()$ evaluates the safety levels of driving behaviors, and its value is the sum of the safety level $\mu$ of each driving behavior in a task plan.
In this work, we separately compute the safety level of each driving behavior, and do not consider their interactions. 
A breach of user preferences $\textit{cst}^\textit{soft}$ incurs a violation cost to \textit{Pref}$()$.
\textit{Cost}$()$, \textit{Pref}$()$, and \textit{Safe}$()$ generate positive real numbers or zeros. 
\begin{equation}\label{eqn:exp_utility}
\small
\begin{aligned}
    \textit{Utility}(p)=\mathbb{E}\big(& \alpha_0\!\cdot\!\textit{Cost}(p)\!+\! \alpha_1\!\cdot\!\textit{Pref}(p)  \!+\! \alpha_2\!\cdot\!\textit{Safe}(p)\big)
\end{aligned}
\end{equation}
where $p$ denotes the generated task plan which is implemented by the computed motion trajectories.
Each trajectory corresponds to one driving behavior, and the trajectories together form one continuous path.
$\alpha_0$ and $\alpha_1$ are negative constants, and $\alpha_2$ is a positive constant. 

\vspace{0.8em}
\noindent\textbf{Format of Algorithm Output:}
The driving agent computes a task-motion plan that includes a sequence of $N$ driving behaviors, and a sequence of $N$ motion trajectories that are head-to-tail connected, where each trajectory corresponds to one driving behavior. 
\emph{The driving agent's goal is to compute task-motion plans towards maximizing the overall expected utility presented in Eqn.~(\ref{eqn:exp_utility}).}

Next, we present our grounded layered planning algorithm for complex urban driving tasks. 


\begin{algorithm}[t]
\small
\caption{Grounded Layered Autonomous Driving}\label{alg:alg_GLAD}
\begin{algorithmic}[1]

\STATE \textbf{Function} \textit{OptimalPlan}($\textit{cst}^\textit{hard}$, $\textit{cst}^\textit{soft}$, $\mu$, \textit{loc})
\begin{ALC@g}

\STATE Sequence POIs to fulfill $\textit{cst}^\textit{hard}$, where each feasible sequence is denoted as $seq$.\label{l:seq_poi}
\STATE Generate all feasible sequences of driving behaviors for each $seq$ using $\textit{Plnr}^t$, where each plan is denoted as $p$.\label{l:seq_action}
\STATE Generate a motion plan for plan $p$ using $\textit{Plnr}^m$ with \textit{loc}.\label{l:compute_motion}
\STATE Calculate \textit{Cost}$(p)$, \textit{Pref}$(p)$, and \textit{Safe}$(p)$.\label{l:utility}
\STATE Compute an optimal plan $p^*$ from all $p$'s based on Eqn.~(\ref{eqn:exp_utility}).\label{l:optimize}
\end{ALC@g}
\STATE \textbf{EndFunction}

\item[]

\REQUIRE$\textit{Plnr}^t$ and $\textit{Plnr}^m$ \\
\hspace{-1.7em}\textbf{Input:} $\textit{cst}^\textit{hard}$, $\textit{cst}^\textit{soft}$, \textit{IM}, and \textit{loc}\\
\STATE Train a vision-based safety estimator (Section~\ref{sec:safety}).\label{l:train_estimator}
\STATE  Initialize $\mu$ of each behavior as $1.0$.\label{l:initialize_mu}
\STATE Call function \textit{OptimalPlan} to compute an optimal plan $p^*$\label{l:optimal_init_plan}
\WHILE {$\textit{cst}^\textit{hard}$ still has driving tasks}\label{l:check_req_start}
\STATE Extract the first driving behavior in $p^*$.\label{l:extract_action}
\STATE Estimate $\mu$ of the behavior using the trained estimator with \textit{IM} based on Eqn.~(\ref{eqn:net}).\label{l:compute_mu}
\STATE Call function \textit{OptimalPlan} to compute an optimal plan $p'$ with the updated $\textit{cst}^\textit{hard}$, $\mu$, and \textit{loc}.\label{l:re-compute}
\IF{$p^*==p'$}\label{l:check_optimal_start}
\STATE Execute the motion trajectories of plan $p^*$.\label{l:execute_traj}
\IF{an POI in $p^*$ is visited}\label{l:check_visited_start}
\STATE Update $\textit{cst}^\textit{hard}$ by removing completed the driving task.\label{l:update_req}
\ENDIF\label{l:check_visited_end}
\STATE Update \textit{loc} and remove the first driving behavior in $p^*$.\label{l:update_position}
\ELSE
\STATE Update the optimal plan $p^*$ with $p'$.\label{l:update_optimal}
\ENDIF\label{l:check_optimal_end}
\ENDWHILE\label{l:check_req_end}
\end{algorithmic}
\end{algorithm}

\begin{figure*}[t]
\vspace{1.5em}
\begin{center}
    \includegraphics[width=1.95\columnwidth]{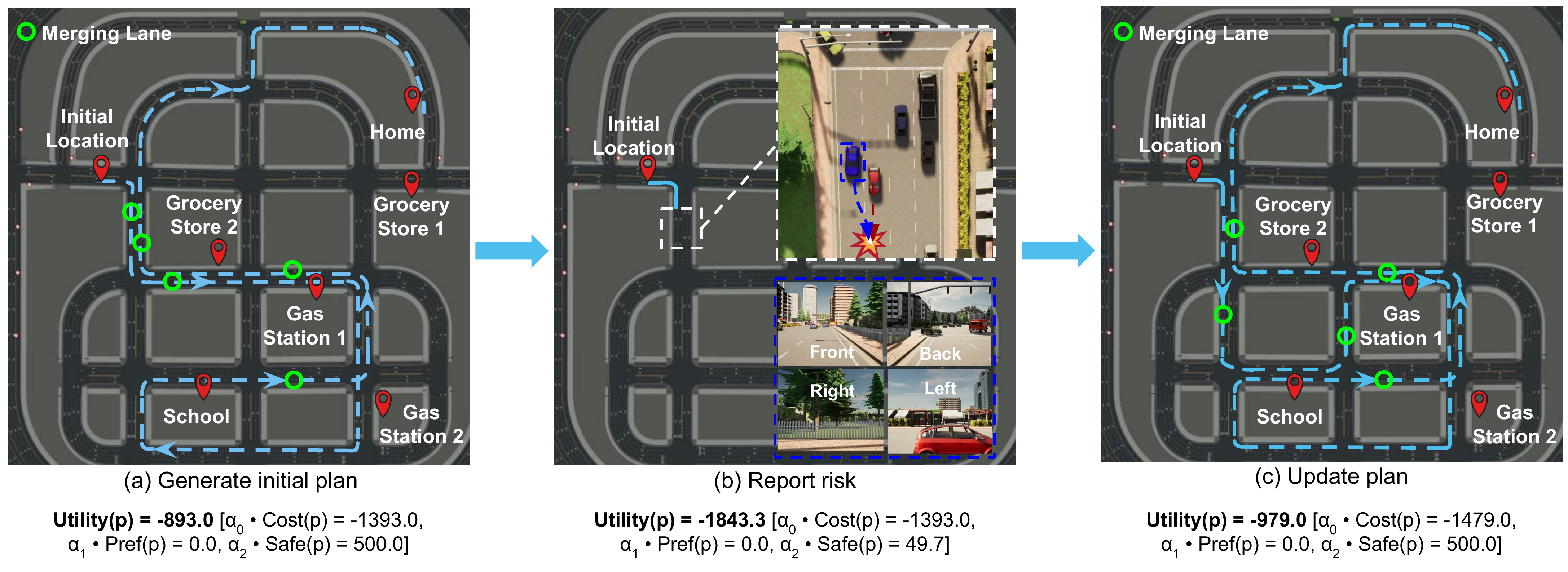}
\end{center}
\vspace{-1em}
\caption{
An illustrative example of GLAD for grounded layered autonomous urban driving. 
The vehicle's \textbf{service task} was to take Emma home after work. 
On the way home, Emma needed to pick up kid from school, stop at a gas station, and visit a grocery store.
To fulfill the service request, our vehicle needed to visit at least four POIs, including School, Grocery Store, Gas Station, and Home. 
~~\textbf{(a)} 
GLAD computed a task-motion plan as shown in \textbf{\color{bleudefrance} blue dashed line}, where at the service level the vehicle planned to visit the following POIs in order: \textit{Gas Station 1}, \textit{School}, \textit{Grocery Store 2}, and \textit{Home}. 
All POIs are marked with \textbf{\color{red} red pins}. 
The planned ``merge lane'' positions are marked with \textbf{\color{green} green circles}. 
~~\textbf{(b)} Our vehicle (\textbf{\color{blue} a blue car}) was preparing to \textbf{merge left} in the highlighted area, 
and observed that there was \textbf{\color{red}a red car} making it unsafe to merge left. 
~~\textbf{(c)} Based on the computed safety value, GLAD generated a new task-motion plan that helped avoid merging lane in the highlighted area. 
Although the new plan required a longer traveling distance, it significantly improved driving safety, while considering user preferences. 
Following the updated plan, the vehicle was able to fulfill the service request. 
}\label{fig:illustration}
\end{figure*}

\section{Algorithm}

In this section, we describe Grounded Layered Autonomous Driving (GLAD), a novel planning algorithm to address the problem of ``planning to complete complex urban driving tasks'', as defined in Section~\ref{sec:problem}, as well as its vision-based safety estimation component.

\subsection{GLAD Algorithm}\label{sec:alg}
Algorithm~\ref{alg:alg_GLAD} presents our GLAD algorithm -- the main contribution of this work. 
The input of GLAD includes a task planning system $\textit{Plnr}^t$, a motion planning system $\textit{Plnr}^m$, a service request $\textit{cst}^\textit{hard}$, user preferences $\textit{cst}^\textit{soft}$, observation $\textit{IM}$, and the current location \textit{loc} of our vehicle.

GLAD begins with training our vision-based safety estimator.
The estimator predicts the safety level $\mu \in [0.0, 1.0]$ of each driving behavior using observation \textit{IM} (\textbf{Line~\ref{l:train_estimator}}).
GLAD initializes the safety level $\mu$ of each driving behavior as $1.0$ (\textbf{Line~\ref{l:initialize_mu}}), which indicates the vehicle optimistically believes all driving behaviors in a plan are absolutely safe.
GLAD then calls the function \textit{OptimalPlan} to compute an optimal task-motion plan $p^*$ (\textbf{Line~\ref{l:optimal_init_plan}}).

In the \textit{while-loop}, the vehicle completes driving tasks in plan $p^*$, while at the same time plan $p^*$ will be continually updated (\textbf{Lines~\ref{l:check_req_start}}$\sim$\textbf{\ref{l:check_req_end}}).
The \textit{outer if-else} iteratively executes the motion trajectory for the vehicle to visit POIs in plan $p^*$ (\textbf{Lines~\ref{l:check_optimal_start}}$\sim$\textbf{\ref{l:check_optimal_end}}).
The \textit{inner if} is used to check if an POI in plan $p^*$ is visited (\textbf{Lines~\ref{l:check_visited_start}}$\sim$\textbf{\ref{l:check_visited_end}}).
At each iteration in the while-loop, the safety level $\mu$ of an driving behavior in plan $p^*$ is estimated using the trained safety estimator with given \textit{IM} based on Eqn.~(\ref{eqn:net}) (\textbf{Line~\ref{l:compute_mu}}).
A new optimal plan $p'$ is then computed with the updated $\mu$ and \textit{loc} (\textbf{Line~\ref{l:re-compute}}).
If $\textit{Plnr}^t$ suggests the same plan (\textbf{Line~\ref{l:check_optimal_start}}), the vehicle will continue to perform the driving behavior at the motion level  (\textbf{Line~\ref{l:execute_traj}}).
Otherwise, the \emph{currently optimal} $p'$ will replace plan $p^*$ (\textbf{Line~\ref{l:update_optimal}}).
After a driving behavior is executed at the motion level, \textit{loc} will be updated (\textbf{Line~\ref{l:update_position}}). 

Next, we discuss how to compute safety level $\mu$ using the developed safety estimator (\textbf{Line~\ref{l:compute_mu}}).

\subsection{Vision-based Safety Estimator}\label{sec:safety}
Safety level $\mu$ of driving behavior is computed using a binary classifier $\Theta$ and an encoder $\Psi$, where the two are separately considered to accommodate different implementations of $\Theta$ in our system. 
\begin{equation}\label{eqn:net}
   \mu = \Theta \big (\Psi(\textit{IM}) \big)
\end{equation}
where $\Theta()$ generates probability $\mu \in [0.0, 1.0]$ -- the safety level of the current driving behavior. 
$\Psi()$ generates abstract features of road conditions, such as states of surrounding cars and distance to the center of the lane, at the time before executing the driving behavior.
Here, we used an off-the-shelf pre-trained encoder from the literature~\cite{toromanoff2020end}, which was implemented using ResNet-18~\cite{he2016deep}.
$\Psi()$ can output a feature vector of the size $32768$.

Fig.~\ref{fig:illustration} illustrates GLAD, where the vehicle found it difficult to execute its current plan due to the traffic (visually perceived). 
Accordingly, GLAD updated the plan to ensure safety and efficiency while respecting user preferences.

\section{Algorithm Instantiation}\label{sec:instantiation}

\noindent \textbf{Task planning system $\textit{Plnr}^t$:} We model seven driving behaviors in ASP~\cite{lifschitz2002answer}, such as \texttt{mergeleft}. 
For instance, action \texttt{mergeleft(L1,L2)} navigates a vehicle from lane \texttt{L1} to lane \texttt{L2}, whose description is shown below:

\begin{footnotesize}
\begin{verbatim}
inlane(L2,I+1) :- mergeleft(I), inlane(L1,I),
                leftof(L2,L1), step(I), I>=0, I<n.    
\end{verbatim}
\end{footnotesize}
where the right side of \verb|:-| describes the action and its \emph{preconditions}, and the left side is the \emph{effect}.
Specifically, its precondition is that the vehicle locates in the lane \texttt{L1} at step \texttt{I} (i.e., \texttt{inlane(L1,I)}), and lane \texttt{L2} is on the left of \texttt{L1} (i.e., \texttt{leftof(L2,L1)}).
The corresponding effect is that the vehicle locates in lane \texttt{L2} at step \texttt{I+1}, i.e., \texttt{inlane(L2,I+1)}.
The lanes and their spatial relationships, e.g., \texttt{leftof()}, could be easily extracted from a given map.
More details about other behaviors are provided in the supplementary materials.

\vspace{0.8em}
\noindent \textbf{Motion planning system $\textit{Plnr}^m$:} 
$\textit{Plnr}^m$ was implemented using the ``BasicAgent'' function provided by CARLA, which navigates our vehicle to a given target destination from its current position, and also enables our vehicle to follow traffic rules and respect other vehicles. 
Specifically, $\textit{Plnr}^m$ is composed of a path planner and a tracking controller.
The $A^*$ algorithm is used to compute optimal motion trajectories.
A proportional-integral-derivative (PID) controller is applied to generate control signals, e.g., for steering, throttle, and brake.

\vspace{0.8em}
\noindent \textbf{Safety Estimator} was implemented in two ways.
One is based on Artificial Neural Networks (ANNs)~\cite{mcculloch1943logical} using PyTorch~\cite{paszke2017automatic}, and the other is based on Support-vector machine (SVM)~\cite{cortes1995support} using Scikit-Learn~\cite{pedregosa2011scikit}. 
The ANN model contains one fully connected hidden layer with $1024$ nodes, and the output layer containing $2$ nodes is connected to the Softmax function.
The negative log-likelihood loss (i.e., NLLLOSS) was applied.
We also implemented SVM for binary classification. 
In this work, we selected Support Vector Classification (SVC), where the parameter ``gamma'' was set as ``auto'' and the function ``predict\_proba'' was enabled to get the probability of two classes labels.
Given $\Psi(\textit{IM}$), the implemented ANN (or SVM) could generate the probability of the current driving behavior being safe (i.e., the safety level $\mu \in [0.0, 1.0]$). 

We used the following parameters of Eqn.~(\ref{eqn:exp_utility}) for plan optimization: $\alpha_0=-1.0$, $\alpha_1=-1.0$, and $\alpha_2=500.0$.

\section{Experiments}
\label{sec:exp}

We use CARLA~\cite{dosovitskiy2017carla}, an open-source 3D urban driving simulator, for demonstration and evaluation in this work. 
CARLA has been widely used for autonomous driving research~\cite{CARLA_Challenge}. 
In particular, we selected Town05 for our experiments, which is featured by its multi-lane roads, and is particularly useful for evaluating complex driving behaviors.

\begin{figure}[t]
\centering
\includegraphics[width=1.0\columnwidth]{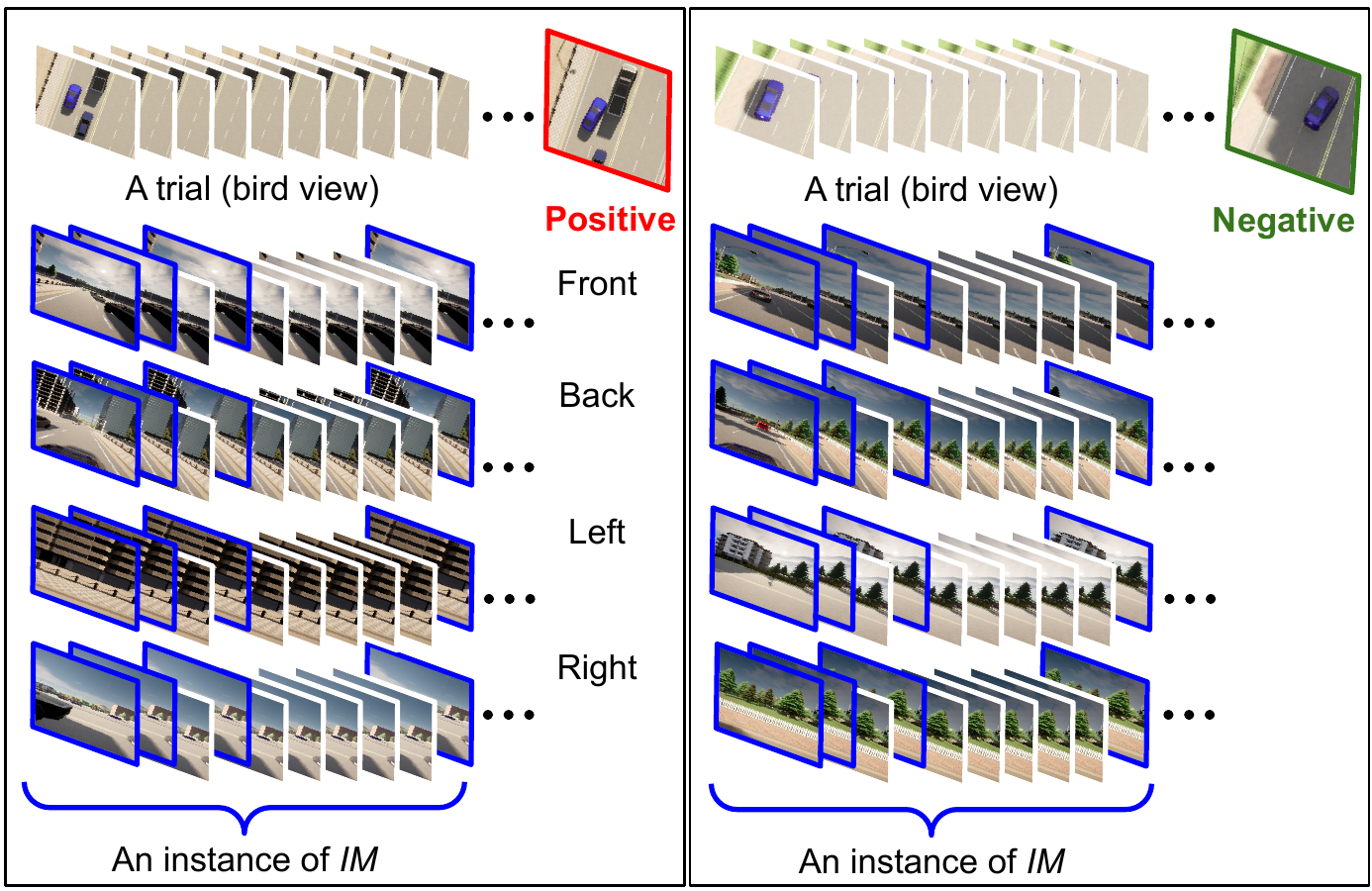}
\vspace{-1.5em}
\caption{Two instances of \textit{IM} with different positive (\textbf{Left}) and negative (\textbf{Right}) labels in the \textit{IM} dataset, where positive and negative indicate merging lane is unsafe and safe, respectively.
}\label{fig:collection}
\end{figure}

\vspace{0.8em}
\noindent \textbf{\textit{IM} Dataset Collection:}
We built an image dataset to train the safety estimator, where the dataset contains $13.8k$ instances. 
Each instance is referred to as \textit{IM}, which includes the \textit{1st}, \textit{2nd}, \textit{4th}, and \textit{10th} frames, as marked in blue box in Fig.~\ref{fig:collection}. 
Right after taking an \textit{IM}, the vehicle was forced to merge left (or right) using a predefined motion trajectory. 
In each trial, the vehicle was given ten seconds to complete the behavior. 
By the end of the ten seconds, the \textit{IM} was labeled \textit{positive}, if there was a collision with a surrounding vehicle, or there existed another vehicle that was very close to our vehicle (the threshold was 1.0 meter). 
Otherwise, the instance was labeled \textit{negative}.
Fig.~\ref{fig:collection} shows two instances (\textit{positive} on the left, and \textit{negative} on the right), where we also present bird views of the two trials. 
Our dataset is also well balanced, as $46.5\%$ of instances are labeled positive, and the remaining ones are labeled negative.
To ensure the diversity of driving scenarios, instances were sampled from 24 different roads. 
The dataset has been \textbf{open-sourced}, where the link for downloading and additional information are provided in the supplementary materials.




\vspace{0.8em}
\noindent \textbf{Experimental Setup:}
The NumPy random seed was fixed as 42.
Seventy percent of instances in dataset \textit{IM} were used for training, and the remaining instances were used for testing. 
The training dataset is denoted as $\textit{D}_\textit{train}$. 
There are two training settings, where the non-default is more challenging. 

\vspace{.2em}
\begin{itemize}[leftmargin=12pt]
    \item \textbf{Default}: We used the same roads for training and testing. 
    \vspace{-1em}
    \item \textbf{Non-default (marked with $\#$}  in figures): We used different roads for training and testing. 
\end{itemize}
\vspace{-.2em}



\subsection{Evaluation Metrics and Three Baselines}
The main measure for evaluation is utility, which considers traveling cost, driving safety, and user preference. 
The cost was evaluated using traveling distance in meter. 
Each collision or unsafe behavior incurred a penalty cost of $15000.0$. 
The violation cost in \textit{Pref}$()$ is $300.0$. 


Three baselines were used in our experiments including:
\begin{itemize}
    \item TPAD: It forced all task-level actions to be executed by the vehicle at the motion level while driving behaviors' safety values are ignored~\cite{chen2015task}. 
    \item TMPUD: It considers motion costs and safety in task-motion planning, but does not model user preferences or visual perceptio~\cite{ding2020task}.
    \item MINI (weakest): It considers safety and user preferences, and minimizes plan length, but it does not compute motion costs. 
\end{itemize}

We use GLAD to refer to the default implementation whose safety estimator was realized using ANN (see Section~\ref{sec:instantiation}). 
By comparison, we use GLAD-SVM to refer to the version that used SVM (for replacing ANN).
The safety estimator implemented by ANN is the default denoted as SE-ANN, and the other is denoted as SE-SVM.


\subsection{Full Simulation}
\label{sec:full}
Experiments conducted in CARLA are referred to as being in \textbf{full simulation}.
We first spawned $\sigma=120$ vehicles with random models and colors in random positions on the map. 
All of them were in autopilot mode, where their speed is less than $30km/h$. 
Then our ego vehicle was spawned at the initial location, as shown in Fig.~\ref{fig:illustration}. 
Each reported value corresponds to an average of 100 trials, where on average, each trial took more than 10 minutes to complete. 
Experiments were performed using a desktop machine (i7 CPU, 32GB memory, and GeForce RTX 3070 GPU).


\vspace{.8em}
\noindent \textbf{Results:}
Table~\ref{tab:real} summarizes the performances of GLAD (ours) and three baselines. 
GLAD achieved the highest overall utility, while at the same time performing the best in respecting user preferences (``\emph{Pref}'') and maintaining safety (``\emph{Safe}''). 
In ``\emph{Cost},'' baseline TPAD produced the best performance, because it does not evaluate execution-time safety and frequently runs into accidents. 


\begin{table}[t]
\vspace{1.5em}
\small
\caption{Utility, cost, penalty caused by violating user preferences and risky behaviors for four algorithms}\label{tab:real}
\centering
\begin{tabular}{cc|c|c|c|c}
\toprule
\multicolumn{2}{c|}{\multirow{2}{*}{Approaches}} & \multirow{2}{*}{\emph{Utility}} & \multirow{2}{*}{\textit{Cost}} & \multirow{2}{*}{\textit{Pref}} & \multirow{2}{*}{\textit{Safe}} \\
\multicolumn{2}{c|}{} & & & &\\ \hline
\multicolumn{2}{c|}{\emph{GLAD (ours)}} & \textbf{-2415.7} & -2265.7 & \textbf{0.0} & \textbf{-150.0} \\ \hline
\multicolumn{2}{c|}{\emph{TMPUD}} & -2569.2 & -1936.2 & -333.0 & -300.0\\ \hline
\multicolumn{2}{c|}{\emph{TPAD}} & -2834.0 & \textbf{-1334.0} & \textbf{0.0} & -1500.0 \\ \hline
\multicolumn{2}{c|}{\emph{MINI}} & -2993.0 & -2543.0 & -300.0 & \textbf{-150.0} \\ 
\bottomrule
\end{tabular}
\end{table}

\begin{figure}
\centering
\vspace{1em}
\includegraphics[width=0.95\columnwidth]{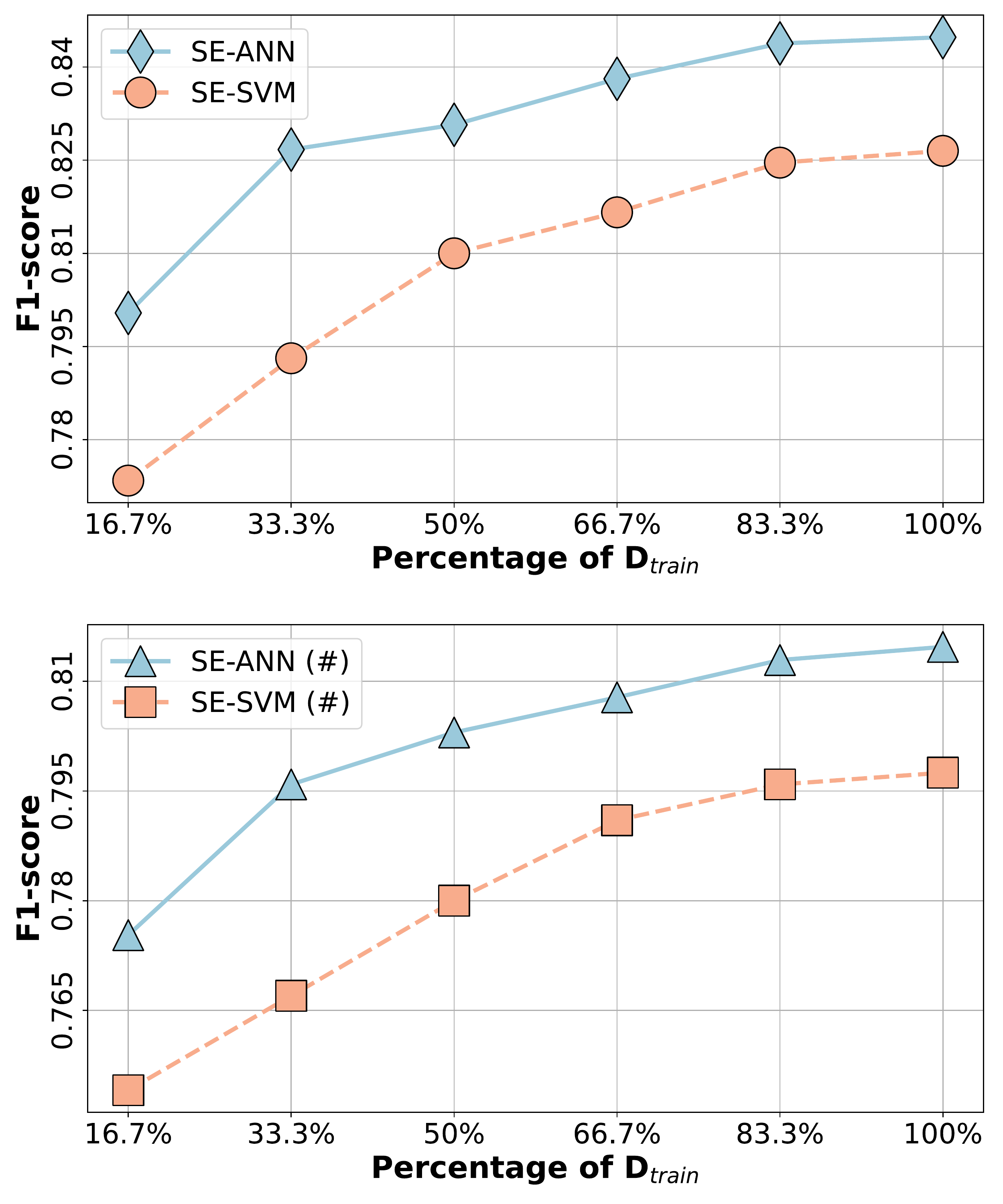}
\vspace{-0.5em}
\caption{Performance of safety estimators SE-ANN and SE-SVM under two training settings: Default (\textbf{Top}), and Non-default (\textbf{Bottom}). 
\textit{x-axis} represents the percentage of $\textit{D}_\textit{Train}$ used in training, and \textit{y-axis} represents the F1-score. 
}\label{fig:safety}
\end{figure}

\vspace{0.8em}
\noindent \textbf{Results from Safety Estimator}:
Fig.~\ref{fig:safety} shows the performances of SE-ANN and SE-SVM in F1-score under two settings (default and non-default). 
From the results, we observe that SE-ANN consistently performed better than SE-SVM. 
Another observation is that safety estimation is more difficult under the non-default setting, where the vehicle used different roads for training and testing. 
Finally, we could see that the performances of both methods increase with more training data.


\subsection{Abstract Simulation}

It is computationally expensive to run experiments in full simulation with CARLA, e.g., running the experiments reported in Section~\ref{sec:full} took more than 72 hours on a modern desktop machine. 
For extensive evaluations, we developed an \textbf{abstract simulation} platform, where the perception component is abstracted. 
Based on the performance reported in Fig.~\ref{fig:safety}, we modeled the vision-based safety estimator using a 2-by-2 confusion matrix. 
In the simulation, we first sampled from the confusion matrix, and then accordingly sampled a safety value $\mu$ based on our safety estimator's performance on our dataset. 
Further, we created two traffic conditions in abstract simulation, heavy traffic and normal traffic, that are associated with different $\lambda$ values, where $\lambda$ is the probability of collisions (or unsafe behaviors). 
In experiments, $\lambda=0.05$ (or $0.08$) under normal (or heavy) traffic. 
Each data point in abstract simulation corresponds to an average of 6400 trials, where we also evaluated their standard deviations. 

\begin{figure}
\centering
\vspace{.5em}
\includegraphics[width=0.95\columnwidth]{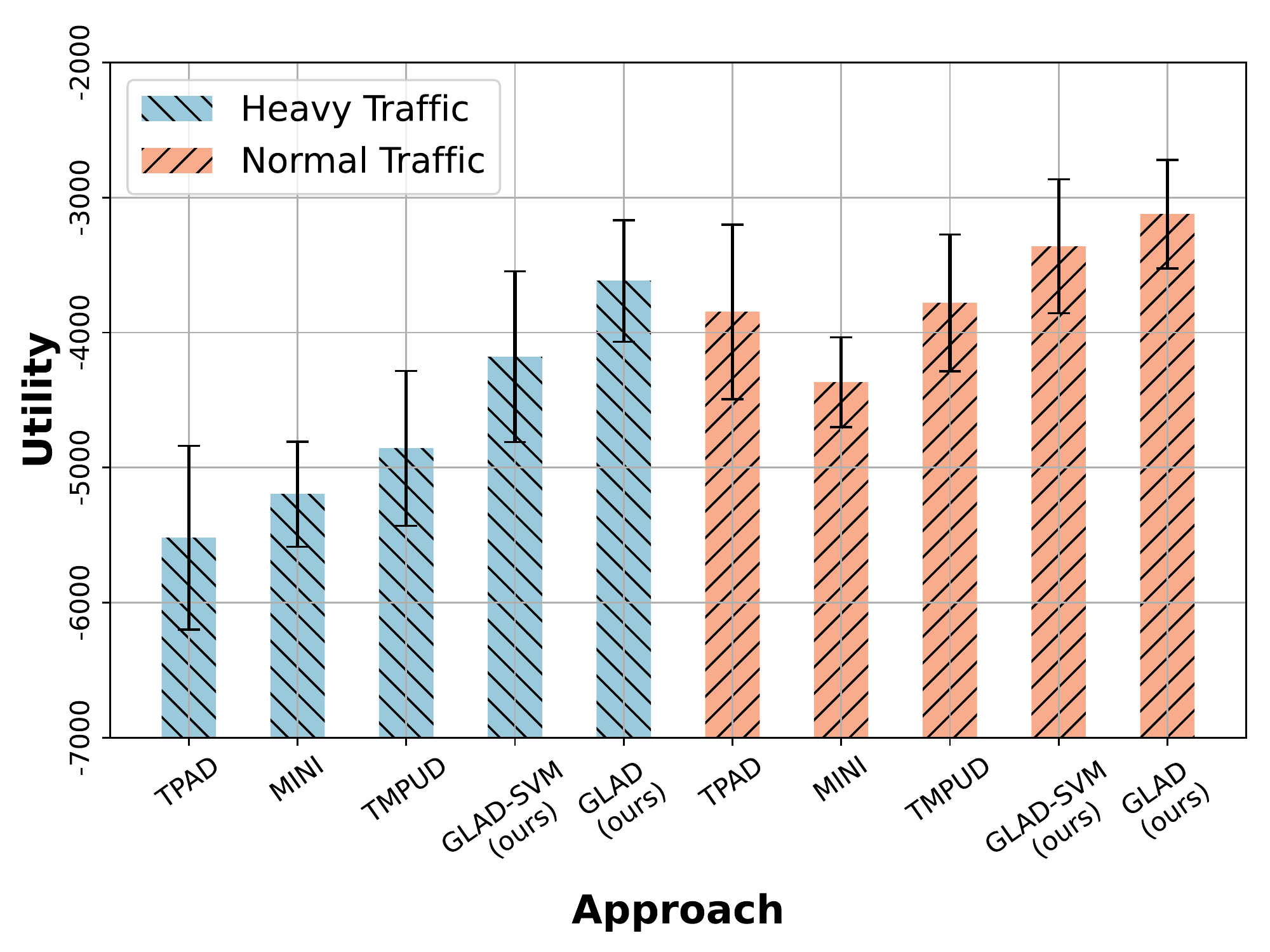}
\vspace{-0.5em}
\caption{Overall performances of GLAD and three baselines, where the \emph{x-axis} represents different methods, and the \emph{y-axis} represents the average utility value. 
We also reported the standard deviations on top of each bar. 
}\label{fig:utility}
\end{figure}

\vspace{0.8em}
\noindent \textbf{Results from Abstract Simulation}:
Fig.~\ref{fig:utility} shows the overall performances of GLAD and the baselines under normal and heavy traffic conditions.
We see that GLAD produced the highest utility under both traffic conditions.
The experimental results demonstrate that GLAD performed the best in maximizing overall utility, which considers task-completion efficiency, road safety, and user preferences.
We also observe that GLAD outperformed the baselines by a larger margin under heavy traffic, because GLAD is good at evaluating execution-time safety to adjust task plans.

\section{Conclusion and Future work}
In this paper, focusing on urban driving scenarios, we develop a vision-based safety estimator and a grounded layered planning algorithm, called GLAD.  
Different from existing urban driving methods, GLAD considers user preference, road safety, and task-completion efficiency -- all at the same time. 
We have extensively evaluated GLAD using a 3D urban driving simulator (CARLA). 
Results suggest that GLAD produced the best performance in overall utility, while maximizing task-completion efficiency, satisfying user preferences, and ensuring the safety of driving behaviors. 

There is room to improve the vision-based safety estimator, where researchers can try other encoders that better capture abstract visual features. 
Another direction for future work is to evaluate the GLAD in challenging road conditions, such as weather, traffic, time, and driving behaviors of surrounding vehicles. 
The contextual information can potentially further improve the overall performance.


\bibliographystyle{IEEEtran}
\bibliography{ref}


\end{document}